\documentclass[dvipsnames,11pt]{article}

\usepackage{acl}

\usepackage{times}
\usepackage{latexsym}

\usepackage[T1]{fontenc}
\usepackage[utf8]{inputenc}

\usepackage{inconsolata}

\usepackage{amssymb}% Nécessaire pour \mathbb
\usepackage{amsmath}

\usepackage{booktabs}
\usepackage[tikz]{bclogo}

\usepackage{tikz}
\usetikzlibrary{arrows.meta}

\usepackage{hyperref}
\usepackage{url}
\usepackage{longtable}
\usepackage{fancyvrb}
\usepackage{multicol}

\usepackage{tcolorbox}
\usepackage{soul}
\usepackage{graphicx}
\usepackage{multirow}
\usepackage{verbatim}

\usepackage{linguex}
\usepackage{hyperref} % Ensure hyperref is included for reference
\usepackage{arydshln}
\usepackage{wasysym}
\usepackage{tikz}
\usetikzlibrary{positioning, shapes.geometric}
\usepackage{xcolor}
\usepackage{listings}
\usepackage{adjustbox}
\usepackage{comment}
\usepackage{mdframed}
\includecomment{comment}

\definecolor{dkgreen}{rgb}{0,0.6,0}
\definecolor{gray}{rgb}{0.5,0.5,0.5}
\definecolor{mauve}{rgb}{0.58,0,0.82}
\definecolor{green1}{HTML}{077500}

\title{DIMSUM: Discourse in Mathematical Reasoning as a Supervision Module}

\begin{document}

\author{
    \textbf{Krish Sharma\textsuperscript{1,2,†}},  
    \textbf{Niyar R Barman\textsuperscript{1,2,†}}, 
    \textbf{Akshay Chaturvedi\textsuperscript{1}}
    ,
    \textbf{Nicholas Asher\textsuperscript{1}}
    \\ 
    \\
    \textsuperscript{1}IRIT, Toulouse, France  
    \\ 
    \textsuperscript{2} NIT Silchar, India  
}
\maketitle

\begingroup
    \renewcommand\thefootnote{\textsuperscript{†}}
    \footnotetext{Equal Contribution}
\endgroup

\begin{abstract}

We look at reasoning on  GSM8k, a dataset of short texts presenting primary school, math problems.  We find, with \citet{gsm-symbolic}, that current LLM progress on the data set may not be explained by better reasoning but by exposure to a broader pretraining data distribution.  We then introduce a novel information source for helping models with less data or inferior training reason better: discourse structure.  We show that discourse structure improves performance for models like Llama2 13b by up to 160\%. Even for models that have most likely memorized the data set, adding discourse structural information to the model still improves predictions and dramatically improves large model performance on out of distribution examples.

\end{abstract}

\section{Introduction}

Recent advancements in large language models (LLMs) have led to impressive performance on reasoning tasks, particularly on benchmark datasets like GSM8K \cite{cobbe2021trainingverifierssolvemath}. These models exhibit strong problem-solving abilities, often producing solutions that resemble human-like reasoning. However, recent studies have demonstrated that altering the entities or numerical values can degrade the reasoning capabilities of LLMs
\cite{gsm-symbolic}.  Other studies have also shown that model reasoning is not robust \cite{schaeffer:etal:2024,valmeekam:etal:2022,asher:bhar:2024,kambhampati:2024} .  This raises a critical question: do these models genuinely engage in reasoning, or do they primarily rely on memorization and pattern recognition? These findings suggest a need for a deeper investigation into how LLMs process logical structures and linguistic variations.

Prior research \cite{chen:etal:2024} has also investigated the effect of permuting the premise order in GSM8K problems, showing that such modifications affect LLMs' reasoning capabilities. This observation suggests that, unlike formal logic, where the validity of a conclusion remains unchanged regardless of premise order, natural language reasoning is shaped by structural dependencies within discourse.

%A linguistic perspective explains phenomenon can be attributed to discourse structure. 
A linguistic perspective explains this dependency.  Unlike formal logic, natural language introduces ambiguity and underspecification, which are resolved through contextual and structural cues that tell us how clauses in a text are semantically related. The ordering of clauses within a discourse, for instance, significantly impacts interpretation, and determines, in the absence of other cues, temporal and causal relationships. Consider the simple sequences in \ref{and} from \cite{asher:1993}:
\ex. \label{and}
\a. Pat took off his shoes. Pat got in bed.
\b. Pat got in bed.  Pat took off his shoes

When we permute the order of the clauses as in \ref{and}b, most native English speakers will draw different conclusions. They will conclude from in \ref{and}a that Pat first took off his shoes before getting into bed, whereas in \ref{and}b that Pat took his shoes off after he got in bed.  Ordering in this case determines a semantic relation--\cite{asher:1993} calls it Narration that entails that the event in the second clause follows the event in the first.

Even in seemingly simple narratives, semantic relations such as Elaboration, Background information, and  Narration shape how reasoning unfolds. Additionally, discourse structure affects anaphora resolution—reordering premises can change how pronouns and referents are interpreted, thereby influencing reasoning outcomes.  For example, in \ref{anaphora} 
\ex. \label{anaphora} 
John picked 3 apples.  Sam picked 4 apples.  He then picked 2 more.

most native speakers would pick Sam as the referent of {\em he} in the last sentence.  If we reverse the first two sentences, most speakers would pick John as the referent of {\em he}.

We show that information about discourse structure improves performance for all LLMs on the GSM8k data set, but especially for older, models like Llama2 13b with poor reasoning performance by up to 160\%. Even for models that have most likely memorized the data set, adding discourse structural information to the model improves performance on GSM8K and dramatically improves performance on out of distribution examples.

\begin{figure*}[ht]
\centering
\begin{tcolorbox}[
  colback=gray!10,         
  colframe=black,          
  width=\textwidth,        
  title=\textbf{\footnotesize \textsc{Example of Discourse Structure:}},  
  fonttitle=\footnotesize, 
  sharp corners,           
  boxsep=4pt,              
  top=6pt,                 
  bottom=6pt               
]
{\footnotesize \ttfamily

% \textcolor{black}{\textbf{\colorbox{Apricot}{System Prompt:}}} \texttt{< System Prompt >}

% \noindent\makebox[\linewidth]{\dotfill}

\textcolor{black}{\textbf{\colorbox{Salmon}{Question:}}}
Oliver picks 44 kiwis on Friday. Then he picks 58 kiwis on Saturday. On Sunday, he picks double the number of kiwis he did on Friday, but five of them \underline{were} a bit smaller than average. How many kiwis does Oliver have? 

\noindent\makebox[\linewidth]{\dotfill}

\textcolor{black}{\textbf{\colorbox{Apricot}{Discourse Structure:}}}\\
\textcolor{black}{\textbf{<Structure>}}

\hspace{1em}\textbf{Topics:} 
\begin{verbatim}
        [topic-a] Oliver’s kiwi picking on consecutive days
        [topic-b] Sunday’s special condition (five smaller kiwis)    
\end{verbatim}

\hspace{1em}\textbf{Relationships:} $t_a \xrightarrow{\text{Elab}} t_b$

\hspace{1em}\textbf{Premises:}
\begin{verbatim}
        P1: Picks 44 kiwis on Friday        P2: Picks 58 on Saturday        P3: On Sunday picks double the 
        Friday amount       P4: Sunday has five smaller kiwis       P5: Oliver wants to know total kiwis
\end{verbatim}

\hspace{1em}\textbf{Narrative Flow:} \\
\begin{center}
    \begin{tikzpicture}[>=Stealth, node distance=2.5cm, on grid, auto]
  % Define nodes
  \node (p1) {\textbf{P1}};
  \node (p2) [right=of p1] {\textbf{P2}};
  \node (p3) [right=of p2] {\textbf{P3}};
  \node (p5) [right=of p3] {\textbf{P5}};
  \node (p4) [below=1.5cm of p3] {\textbf{P4}};

  % Draw arrows
  \draw[->] (p1) -- node [above] {narr} (p2);
  \draw[->] (p2) -- node [above] {narr} (p3);
  \draw[->] (p3) -- node [above] {narr} (p5);
  \draw[->] (p3) -- node [left] {bckgnd} (p4);
\end{tikzpicture}
\end{center}

% \[
% \textbf{P1}
%   \xrightarrow{\text{Elab}}
% \textbf{P2}
%   \xrightarrow{\text{Narr}}
% \textbf{P3}
%   \xrightarrow{\text{Narr}}
% \textbf{P4}
%   \xrightarrow{\text{Narr}}
% \textbf{P5}
% \]
\textcolor{black}{\textbf{</Structure>}}

} 

\end{tcolorbox}
\caption{Overview of the prompt sequence on an example from \cite{gsm-symbolic} that GPT-o1-mini and Llama-3-8B couldn't solve.  With this prompt all models tested solved the problem correctly (see Appendix~\ref{app:kiwi-example} for model generation outputs).  The full structure generation prompt (\ref{fig:discourse-prompt}), answer generation prompt (\ref{fig:answer-prompt}) and few-shot examples (\ref{app:few-shot-examples}) are available in the appendix .}
\label{fig:thomas_bills}
\end{figure*}

\pagestyle{plain}

\section{Hard GSM8K Dataset and Variants}  
\label{section3}

To systematically assess whether LLMs really reason or just rely on memorized patterns, we use the more difficult portion of the GSM8K data set, {\em Hard GSM8K}, (with problem statements involving 4 or more premises).  We also provide a controlled variant of Hard GSM8K to create out-of-distribution (OOD) variants using the transformations defined below (see Appendix \ref{app:gsm-mod-ex} for examples). This data will be made publicly available. Unlike the data in \citet{gsm-symbolic}, where modifications involves changing entity names and  adding or subtracting premises, our variants are more complex in terms of contextual modification and number of substitutions. We investigate whether the model can maintain performance under variations in numerical values and contextual framing. This data enables us to rigorously evaluate the robustness of LLMs' reasoning, and it serves as a test-bed to study the influence of the discourse structure.

%  Our approach preserves the underlying computational structure but alters key components such as entities (objects or people involved) and actions (events or operations performed), ensuring a controlled shift in problem formulation. % We use the following transformations to shift problem statements out of distribution:  

% \hidden{
% A farmer is selling ducks and chickens at the market. He sells ducks for $10 and chickens for $8. He sells 5 chickens and some ducks. 
% A toy store owner is selling action figures and dolls. He sells action figures for $10 and dolls for $8. He sells 5 dolls and some action figures.
% A farmer is selling ducks and chickens at the market. He sells ducks for $12 and chickens for $6. He sells 7 chickens and some ducks.
% A poultry vendor is offering geese and hens at the bazaar. He prices geese at $10 and hens at $8. He sells 5 hens and some geese.

% }

\subsection{Contextual Modification (C-MOD)}  
In this transformation, we alter the real-world context of a problem while preserving its structural framework and numerical relationships, ensuring the modified problem shifts out of the pretraining data distribution. Given a problem statement \( P \), we define a contextual mapping function \( f_C: P \rightarrow P' \), where entities and actions are replaced by semantically distinct counterparts from an unrelated domain. Formally, if \( E = \{e_1, e_2, \dots, e_n\} \) represents entities in the original problem and \( A = \{a_1, a_2, \dots, a_m\} \) denotes actions, then:
\begin{equation}
    f_C(E, A) = (E', A') \quad 
    \text{such that } 
    \begin{aligned}
        &E' \cap E = \emptyset, \\
        &A' \cap A = \emptyset.
    \end{aligned}
\end{equation}
For instance food items are substituted with digital files, and weight was mapped to storage size. 

% \subsection{Verb/Action Variation}  
% This transformation alters the verbs and actions within the problem while keeping the numerical structure unchanged. Given a problem statement \( P \) with an action set \( A \), we define an action modification function \( f_V: A \rightarrow A' \), where:
% \begin{equation}
%     \begin{aligned}
%         \forall a_i \in A, \quad &f_v(a_i) = a'_i, \\
%         &\text{such that} \quad Type(a_i) = Type(a'_i).
%     \end{aligned}
% \end{equation}
% Here, \(\text{Type}(a)\) ensures that the new action remains semantically valid. For instance, in a chemistry variant, "adding brownies" is changed to "mixing a chemical compound," but the computation steps remain identical.

\subsection{Numerical Modification (N-MOD)}  
This transformation alters numerical values while preserving the problem’s reasoning structure. Given a set of numerical values \( N = \{n_1, n_2, \dots, n_k\} \) in a problem, we apply a function \( f_N: N \rightarrow N' \) such that:
\[
\begin{aligned}
    \forall n_i \in N, \quad &f_N(n_i) = n'_i, \\
    &\text{where} \quad \text{Scale}(N') = \text{Scale}(N).
\end{aligned}
\]
  
Here, \(\text{Scale}(N)\) ensures proportionality is maintained. For instance, if the weight of an object triples, the same transformation is applied to the modified numbers.

\subsection{Lexical Modification (L-MOD)}  
This technique modifies words and phrases without altering sentence structure or meaning. Given a vocabulary set \( V = \{v_1, v_2, \dots, v_p\} \), we define a substitution function \( f_L: V \rightarrow V' \) such that:
\[
\forall v_i \in V, \quad f_L(v_i) = v'_i, \quad \text{where } \text{Syn}(v_i) = v'_i.
\]  
Here, \( \text{Syn}(v) \) ensures \( v' \) is a valid synonym or equivalent phrase. This tests whether models are robust to surface-level changes.

\section{Generating discourse structure}

%Discourse structure refers to the systematic organization of text, where the arrangement of sentences encodes key properties of the narrative. In the context of logical reasoning tasks, inconsistencies in textual organization can obscure the underlying logical flow, making it challenging for models to recognize patterns and deduce valid conclusions.

To annotate discourse structures systematically within Hard GSM8K, we identify 10 complex examples containing multiple events and subjects, with more than eight sentences.  We manually annotated these examples according to rules in \cite{asher:lascarides:2003}, used them as few-shot \cite{brown2020languagemodelsfewshotlearners} exemplars for generating annotations across the entire dataset using the Llama 3.1 70B model.

\subsection{Generalized Rules for Annotating Discourse Structure}

Understanding discourse structure requires a series of steps: identifying the main event, state or concept in each premise, defining the relationships between these via the semantic dependencies between the premises.  Inspired by \citet{asher:lascarides:2003}, we offer a set of informal rules in the system prompt to help models build a discourse structure, incorporating temporal markers as a key factor. %We use these to guide models to build a discourse structure.
\subsection{Identifying Eventualities}  

Each premise in our data set is a sentence or clause; and in a discourse structure each one introduces a central {\em eventuality}, an event, state or concept \cite{asher:lascarides:2003}.  The appendix provides some guidelines in this regard \ref{app:finding-events}
%\textbf{Problem Statement:}  
%Thomas withdraws \$1000 in 20-dollar bills. He loses 10 bills on his way home. He spends half of the remaining bills and later triples his money. Finally, he converts everything into 5-dollar bills.  
%\textbf{Identified Topics:}  
- $E_{P1}$: kiwis picked on Friday.  
- $E_{P2}$: those picked on Saturday.  
- $E_{P3}$: kiwis picked on Sunday  
- $E_{P4}$: smaller kiwis on Sunday  
- $E_{P5}$: how many kiwis in total. 

\subsection{Relations Between Clauses and eventualities}  

Once eventualities are identified, their relationships are fixed by various discourse relations like Elaboration, Narration or Background that hold between the premises that express them.  The following glosses help specify the meanings of these relations and how to infer them.\\
%\noindent Specific discourse constraints govern these relationships:
\noindent
\textbf{Narration (Narr$(P_i,P_j)$)}– When $P_j$ introduces an eventuality that sequentially follows the one introduced by $P_i$.  Narration can be established with explicit temporal markers in $P_i$ and $P_j$ ({\em then , next, after that, subsequently, finally,} or explicit sequentially related times like consecutive days of the week or months of the year) % in time or introduces an independent but logically sequential action. Additionally, narration is established 
    or when $P_i$ and $P_j$ form part of a chain of actions leading to the final computation.

    \noindent
\textbf{Elaboration (Elab($P_i,P_j$))}– When $P_j$ provides additional details about the event in $P_i$, making it clearer or more informative. Elaboration also occurs when $P_j$ introduces an intermediate computation step necessary for understanding $P_i$. %$P_i$%, as seen in $E_{a}$ $\rightarrow E_b$.
 %On the other hand, P2 functions as an Elaboration of P1 when it provides additional details about the event in P1, making it clearer or more informative. Elaboration also occurs when P2 introduces an intermediate computation step necessary for understanding P1. This relationship is often triggered by causal or explanatory markers such as because, due to, in order to, and consequently.  
\begin{table*}[htbp]
    \centering
    \fontsize{9pt}{11pt}\selectfont  % Set font size explicitly
    \renewcommand{\arraystretch}{1.2} 
    \setlength{\tabcolsep}{6pt} % Adjust column spacing
    \resizebox{\textwidth}{!}{
    \begin{tabular}{lcccccccccccccc}
        \toprule
         &  &  & \multicolumn{4}{c}{\textbf{GSM-Symbolic}} & \multicolumn{8}{c}{\textbf{GSM-MOD (Ours)}}   
        \\
        \textbf{Model} 
        
          & \multicolumn{2}{c}{\textbf{Hd-GSM8K}} 
          & \multicolumn{2}{c}{\textbf{P1}} 
          & \multicolumn{2}{c}{\textbf{P2}} 
          & \multicolumn{2}{c}{\textbf{C-MOD}} 
          & \multicolumn{2}{c}{\textbf{N-MOD}}
          & \multicolumn{2}{c}{\textbf{L-MOD}}
          & \multicolumn{2}{c}{\textbf{Overall}} \\
        \cmidrule(lr){2-15}
          & DS- & DS+  
          & DS- & DS+  
          & DS- & DS+  
          & DS- & DS+  
          & DS- & DS+  
          & DS- & DS+  
          & DS- & DS+ \\
        \midrule
        Llama 3.1-8B  & 79.2 & 88.2 & 58.2 & 79.4 & 34.6 & 70.2 & 32.3 & 71.7 & 73.7 & 77.8 & 56.6 & 75.7 & 54.2 & 75.1 \\
        Llama 3-8B    & 70.2 & 79.8 & 52.8 & 73.6 & 28.3 & 62.4 & 24.2 & 62.6 & 64.6 & 71.7 & 52.5 & 69.7 & 47.1 & 68.0 \\
        Llama 2-13B   & 18.4 & 48.2 & 8.2  & 34.4 & 6.8  & 34.2 & 9.1  & 39.4 & 12.1 & 41.4 & 10.1 & 42.4 & 10.4 & 41.1 \\
        Llama 3.1-70B & 96.0 & 98.3 & 86.8 & 93.3 & 85.1 & 90.2 & 59.6 & 89.9 & 94.9 & 97.0 & 80.8 & 91.9 & 78.4 & 92.9 \\
        \bottomrule
    \end{tabular}
    }
    \caption{Accuracies comparison of different Llama models on Hard GSM8K, \citet{gsm-symbolic}'s P1 and P2 datasets and our perturbation variants (C-MOD, N-MOD, L-MOD) . DS- = without discourse structure ; DS+ = with discourse structure. Overall score siginfies average of all variants of GSM-MOD. The models with discourse structure (see Figure \ref{fig:thomas_bills}) perform considerably better than their counterpart.}
    \label{table:performance}
\end{table*}

\noindent
 \textbf{Background (Bckgd$(P_i,P_j)$)}– When $P_j$ provides contextual information about $P_i$ that is not directly required for computation but helps in understanding the problem. This is commonly the case when $P_j$ describes an event that happened before $P_i$ but does not impact the reasoning process. Markers such as on {\em Monday, previously, before, in the past} indicate background relationships.  Shift from an active verb sequence to a stative verb (underlined in Figure \ref{fig:thomas_bills}) can also signal Background.  For example, consider the problem statement from Figure \ref{fig:thomas_bills}; P4 is attached with Background to P3, and P5 attaches to P3 with Narration.  The attachment reinforces the constraint that the background information is not to be used in the main computation.

\subsection{Markers for positions in a narrative}

 Discourse structure also exploits markers that tell us about the order of clauses in a text.  The simplest consists of an ordinal enumerations as in {\em first, P1, second P2, third P2, fourth P4}.  In addition, there are markers for first and last clauses: {\em initially, finally}.  When clauses containing these markers appear in positions that contradict their intended sequence, it becomes necessary to reorder the premises to align with the canonical narrative flow. We apply this principle to the examples in \citet{chen:etal:2024} (unfortunately they did not release the whole dataset).  
 
\section{Results}

We evaluated several Llama models (\citet{llama2} \citet{llama3}) --Llama 3.1 70B, Llama 3.1 8B, Llama 3 8B, and Llama 2 13B--on reasoning with Hard GSM8K data and variants. with and without incorporating discourse structural information. We tested models on this data with a four-shot learning setting.  We also tested our approach on  \citet{gsm-symbolic}'s data sets with additional premises that were needed (their P1 and P2). To replicate their method, we used eight shot learning. Table~\ref{table:performance} shows our results.

Table~\ref{table:performance} clearly shows that when models have access to the discourse structure of the problems, their performance improves substantially on the original GSM8K problems, on the variants that we have constructed, and on \citet{gsm-symbolic}'s data sets.  This held true both for large models, for which it is pretty clear that they have seen GSM8k problems in their training, and for older smaller models, whose performance drastically improved with access to discourse structure.  Adding discourse structure allows the poorest performing models catch up somewhat to the better ones.  Table \ref{table:performance} also shows that training regimes affect predictions; Llama 3 and 3.1 share largely the same architecture but have quite different training regimes and make significantly different predictions.

Even more noteworthy is the fact that on our altered, unseen examples, even large models like Llama3.1 70b saw their prediction success rate drop, by almost 40 percentage points for the contextually shifted stories.  Once those models had access to the discourse structure their performance improved by sometimes up to 30 percentage points.  From the table, it is clear that the contextually shifted stories were much more challenging than the other variants we made.  

We also tested models on examples with background information from \citet{gsm-symbolic}, and all models correctly avoided using the background information in the computation. In Appendix \ref{app:r-gsm}, we also show that the Llama 3.1 70B with discourse structure are able to predict the correct answers to all the examples provided in \citet{chen:etal:2024}.

%The incorporation of structural information significantly enhanced the reasoning capabilities of the models. The performance improvements are summarized in Table~\ref{tab:structured_performance}. 

%\begin{table}[h]    \centering
%    \begin{tabular}{l c}
%        \hline
  %      \textbf{Model} & \textbf{Accuracy (\%)} \\
%        \hline
%        LLaMA 3.1 8B   & 88.4 \\
%        LLaMA 3.1 70B  &  \\
%        LLaMA 3 8B     & 79.2 \\
%        LLaMA 2 13B    & 48.4 \\
%        \hline
%    \end{tabular}
%    \caption{Accuracy of various LLaMA models on GSM8K under few-shot training with structural information.}
    \label{tab:structured_performance}
%\end{table}

%\section{Experiments}

\section{Conclusion}
We have shown that providing discourse structure, which gives the semantic relations between premises or clauses in a text, improves the performance of language models in solving problems like those in the GSM8K dataset.  We also show that even models that excel on the standard dataset have substantially worse performance when tested on unseen, out of distribution variants; providing those models with discourse structure dramatically improves their performance on the variants.

\section*{Limitations}

One limitation of our paper is that the smaller models do not generate the discourse structure themselves.  Llama70b provides the discourse structure.  We think that models in the 8B range would need fine tuning to learn such structures \cite{thompson:etal:2024b}. We plan to use  Llama70b as an annotator so that the smaller models can be fine tuned to provide this structure on problem sets like GSM8k. 

We did not test our system on GSM-IC \cite{shi:irrelevant}, because upon inspection many of the problems in the data set involve the addition of {\em irrelevant} information.  Consider this example from GSM-IC for instance.

\ex. Lucy has \$65 in the bank. She made a \$15 deposit and then followed by a \$4 withdrawal. The shoe size of Lucy's brother is 80. What is Lucy's bank balance?

The brother's shoe size really is irrelevant to anything in the story, and including it actually makes the text kind of incoherent at least very awkward.  From a discourse structure perspective, the sentence about shoe size intuitively isn't attached to any other clause in the story. Background is different; Background information is relevant but it is like stage setting or additional information about some object or event that is part of the main narrative.  Our model is trained to find Backgrounds but not irrelevant information.  

\section*{Ethics Statement}
We have demonstrated that leveraging discourse structure can improve performance, although it remains imperfect. This work is grounded in mathematical reasoning, which is a prerequisite for downstream real-world applications involving human–robot collaboration. A lack of numerical understanding in a model can lead to tangible consequences, as humans often rely on numerical expressions when providing instructions to robots. Our findings with variants of the \texttt{gsm-mod} model indicate that these models may rely heavily on memorized data, as further evidenced by \texttt{gsm-symbolic}. Consequently, their reported performance metrics may be inflated, potentially creating a misleading sense of readiness for deployment in real-world scenarios.

%\section*{Acknowledgments}

\bibliography{custom}

\appendix
\clearpage
\section{Appendix}
\label{sec:appendix}
In this appendix, we provide additional details to the main text, including:
\begin{itemize}
    \item \ref{app:finding-events} Guidelines for Finding Eventualities
    \item \ref{app:exp-setup} Experimental Setup
    \item \ref{app:sys-prompts} System Prompts 
    \item \ref{app:few-shot-examples} Examples of Answer Generation
    \item \ref{app:gsm-mod-ex} GSM-MOD Example
\end{itemize}

\subsection{Guidelines for Finding Eventualities}
\label{app:finding-events}

\begin{itemize}
    \item \textbf{Main Subject:} The entity or concept driving the problem statement (e.g., a person making a purchase, a worker repairing equipment).
    \item \textbf{Key Actions:} Events that modify the subject’s state (e.g., buying, losing, doubling, converting).
    \item \textbf{Transitions Between Events:} Phrases like \textit{then, after that, subsequently} indicate topic shifts.
    \item \textbf{Supporting Context:} Background information that does not directly influence the problem’s computation.
\end{itemize}

\subsection{Experimental Setup}
\label{app:exp-setup}

All experiments were conducted at the inference stage using an A100 GPU as the primary computing resource. Each experiment was performed as a single run, utilizing approximately \texttt{600} GPU hours. The experiments were implemented using the \texttt{transformers} library.
\\

\noindent The hyperparameters and settings for answer generation are as follows:

\begin{verbatim}
max_new_tokens: 400
temperature: 0.4
top_p: 0.9
tokenizer.pad_token_id: 18610
tokenizer.padding_side: 'right'
\end{verbatim}

\subsection{System Prompts}
\label{app:sys-prompts}
The system prompts used to generate both the discourse structure (\ref{fig:discourse-prompt}) and the final answers (\ref{fig:answer-prompt}) to each query are below.
\subsubsection{Prompt for structure generation}
\label{fig:discourse-prompt}
\begin{tt}
    Task: Analyze a short story using a structured relational
    framework,ensuring proper sequencing and relational mapping.
    
    Instructions:
    
        Identify Topics and Premises:
            - Assign meaningful topic labels to key elements of the story 
              (e.g., [topic-a]: Initial context, [topic-b]: Character’s key action).
            - List premises (P1, P2, ..., Pn) capturing essential events or actions.
    
        Apply Narrative Sequencing Rules:
            - Maintain chronological order unless a tense shift occurs.
            - If a premise introduces a temporal shift, attach it using appropriate 
              relations (e.g., Narr for sequential events, Bkg for background details).
            - Adjust premise order based on discourse markers and time references.
    
            Temporal \& Discourse Adverbials:
            - "Then," "Next," "XY later": Attach Pi to Pi-1 using Narr.
            - "XY before": Attach Pi to Pi-1 using Narr for reverse ordering.
            - "While": Use Elaboration (elab) to connect related events.
            - "Finally": Ensure Pi is the last premise.
            - "First": Ensure Pi is the initial premise.
    
            Day-Specific Rules:
            - If Pi: "Day n B" and Pi+1: "Day n+1 B," attach Pi to Pi+1 using Narr.
            - If days are out of order, shuffle until proper sequential flow is restored.
    
        Label Relationships Between Elements:
            - Use t1--relation--t2 for topic-level relationships (e.g., cause-effect).
            - Use PX--relation--PY for premise-level relationships.
    
    Output Format:
    
        Topics:
        - List identified topics with brief descriptions.
    
        Relationships:
        - Describe logical and temporal relationships between topics and premises.
    
        Premises:
        - Present premises (P1, P2, ..., Pn) in a logically ordered sequence.
    
        Narrative Structure:
        - Show premise connections based on sequencing rules.
    
        Exclusions:
        - Do not include resolution, calculations, or final answers.
\end{tt}

\subsubsection{Prompt for answer generation}
\begin{tt}
    Task: Given a set of premises and an abstract that defines their properties, 
    determine the answer to the question using only the information provided in 
    the abstract. The abstract provides a structured relational framework, 
    ensuring logical consistency in reasoning.
    
    Instructions:
    
        Identify Topics and Premises:
            - Assign meaningful topic labels to the key elements of the problem 
              (e.g., [topic-a]: Initial context, [topic-b]: Events influencing 
              the outcome).
            - List the premises (P1, P2, ..., Pn), ensuring they contain all 
              necessary descriptive statements.
    
        Apply Narrative Sequencing Rules Between Premises:
            - Maintain chronological order unless explicitly defined otherwise in 
              the abstract.
            - If a premise introduces a temporal or logical shift, attach it 
              accordingly using appropriate relations (e.g., Narr for sequential 
              events, Bkg for background information).
            - Resolve premise reordering constraints based on given discourse markers.
    
        Use the Abstract’s Structured Information to Derive the Answer:
            - Follow the relationships and premises as structured in the abstract.
            - Do not introduce external knowledge; rely solely on the provided 
              abstract and premises.
            - If numerical reasoning is required, apply logical derivations based 
              on the premises.
    
    Output Format:
    
        <answer>
        - Provide the final computed or inferred answer.
        - Ensure it aligns with the logical structure in the abstract.
        </answer>
\end{tt}
\label{fig:answer-prompt}

\subsection{Examples of Answer Generation}
\label{app:few-shot-examples}
\subsubsection{GSM-MOD}
\textbf{C-MOD}: \\
\noindent \textbf{Original Question:}Ken created a care package to send to his brother, who was away at boarding school.  Ken placed a box on a scale, and then he poured into the box enough jelly beans to bring the weight to 2 pounds.  Then, he added enough brownies to cause the weight to triple.  Next, he added  to the package gummy bears whose weight equaled 1/2 of the weight of the package with the brownies and the jelly beans.  And finally, he added enough gummy worms to double the weight once again.  What was the final weight of the box of goodies, in pounds?

\noindent \textbf{Modified Question:} Ken created a package  to send to his brother help him repair his house.  Ken placed a box on a scale, and then he poured into the box enough nails to bring the weight to 2 pounds.  Then, he added enough screws and bolts to cause the weight to triple.  Next, he added  to the package a set of socket wrenches whose total weight equaled 1/2 of the weight of the package so far.  And finally, he added a bunch of drills and bits to double the weight once again.

\noindent \textbf{Structure:}

\begin{tt}
\noindent<structure>

\noindent Topics
[topic-a]: Ken’s package creation for his brother’s house repairs
[topic-b]: Incremental weight additions (nails, screws, bolts, socket
wrenches, drills)

\noindent Relationships
t1--elab--t2

\noindent Premises
P1: Ken created a package to send to his brother to help repair 
his house.
P2: He placed a box on a scale and poured in nails until the weight
reached 2 pounds.
P3: He then added enough screws and bolts to cause the weight to 
triple.
P4: Next, he added a set of socket wrenches weighing half the package’s
current total.
P5: Finally, he added drills and bits to double the weight once again.

\noindent Narrative Structure
P1--narr--P2
P2--narr--P3
P3--narr--P4
P4--narr--P5

\noindent</structure>

\end{tt}

\noindent \textbf{Answer without DS (generated by Llama 3.1 70B)}:

\begin{tt}
    \noindent Step 1: Determine the initial weight of the nails.
    The initial weight of the nails is 2 pounds.

    \noindent Step 2: Calculate the weight after adding screws and bolts.
    After adding screws and bolts, the weight triples. So, the new weight is 2 * 3 = 6
    pounds.
    
    \noindent Step 3: Calculate the weight after adding the socket wrenches.
    The weight of the socket wrenches is 1/2 of the current weight. So, the weight of
    the socket wrenches is 6 / 2 = 3 pounds. The total weight now is 6 + 3 = 9 pounds.
    
    \noindent Step 4: Calculate the weight after adding the drills and bits.
    The weight doubles after adding the drills and bits. So, the new weight is 9 * 2 =
    18 pounds. \textcolor{red}{However, we know the final weight is 42 pounds, not 18 pounds, which}
    \textcolor{red}{means our calculation approach needs adjustment to find the weight of the drills}
    \textcolor{red}{and bits}
    
    \noindent Step 5: Correct the approach to find the weight of the drills and bits.
    Given that the final weight of the package is 42 pounds and this is after the 
    weight has doubled from the addition of the drills and bits, we can work 
    backwards. If the weight doubles to 42 pounds, then before adding the drills and
    bits, the weight was 42 / 2 = 21 pounds.
    
    \noindent Step 6: Calculate the weight of the drills and bits.
    Since the weight before adding the drills and bits was 21 pounds, and after adding
    them it became 42 pounds, the weight of the drills and bits is 42 - 21 = 21 pounds.
    
    \noindent The final answer is: 21
\end{tt}

\noindent \textbf{Answer with DS (generated by Llama 3.1 70B)}
\begin{tt}
    \noindent Step 1: Identify the initial weight of the package
    The initial weight of the package after adding nails is 2 pounds, as stated in
    premise P2.

    \noindent Step 2: Calculate the weight after adding screws and bolts
    The weight triples after adding screws and bolts, so the new weight is 2 * 3 = 6
    pounds, as stated in premise P3.

    \noindent Step 3: Calculate the weight after adding socket wrenches
    The weight of the socket wrenches is 1/2 of the current weight of the package, which is 6 pounds. So, the weight of the socket wrenches is 6 / 2 = 3 pounds. The
    total weight now is 6 + 3 = 9 pounds, as stated in premise P4.

    \noindent Step 4: Calculate the final weight after adding drills and bits
    The weight doubles after adding drills and bits, so the final weight is 9 * 2 = 18
    pounds, as stated in premise P5.

    \noindent The final answer is: 18
\end{tt}
\\
\textbf{N-MOD}:
\\
\textbf{Original Question:}
Sanjay saw a 40-foot dolphin with 12 16-inch remoras attached to it. But a quarter of the remoras go away. What percentage of the dolphin's body length is the combined length of the remaining remoras?
\\
\textbf{Modified Question:}
Sanjay saw a 60-foot dolphin with 16 12-inch remoras attached to it. But a quarter of the remoras go away. What percentage of the dolphin's body length is the combined length of the remaining remoras?

\textbf{Structure:}
\begin{tt}

\noindent <structure>

\noindent Topics
[topic-a]: Sanjay's observation of the dolphin
[topic-b]: The remoras attached to the dolphin

\noindent Relationships
t1--elab--t2

\noindent Premises
P1: Sanjay saw a 60-foot dolphin with 16 12-inch
remoras attached to it.
P2: But a quarter of the remoras go away.

\noindent Narrative Structure
P1--narr--P2

\noindent</structure>
\end{tt}

\textbf{Answer without DS:}
\begin{tt}
First, find the total number of remoras remaining: 16 - 16*1/4 =  12.
Then, find the combined length of the remoras in inches: 12 inches/remora * 12 remoras = 144 inches.
Then divide that number by 12 to convert it to feet: 192 inches / 12 inches/foot = 16 foot.
Then divide the combined remora length in feet by the dolphin's length and multiply by 100\% to express the answer as a percentage: 16 foot / 60 feet * 100\% = 26\%

\end{tt}
\textbf{Answer with DS:}
\begin{tt}
<answer>

From P1, we know that Sanjay saw a 60-foot dolphin with 16 12-inch remoras attached to it. From P2, we know that a quarter of the remoras go away.

Let's first calculate the length of the remaining remoras. If a quarter of the remoras go away, then the number of remoras remaining is:

16 - (16 / 4) = 16 - 4 = 12

So, there are 12 remaining remoras. Their combined length is:

12 * 12 = 144 inches

Now, let's calculate the percentage of the dolphin's body length that the combined length of the remaining remoras is. We know the dolphin is 60 feet long, which is equal to 720 inches (60 * 12).

The percentage of the dolphin's body length that the combined length of the remaining remoras is:

(144 / 720) * 100% = 20%

Therefore, the combined length of the remaining remoras is 20\% of the dolphin's body length.

</answer>

\end{tt}

\subsubsection{GSM-Symbolic-NoOp}
\label{app:kiwi-example}
The example in this subsection is taken from the Symbolic-NoOP variation of \cite{gsm-symbolic}, where strong reasoning models like o1-mini\cite{openai-o1} fail without discourse structure. However, with the correct discourse structure, all models in the Llama family, including Llama 2 13B, produce the correct answer.The highlighted text in answer generation demonstrates how it corrects its reasoning.  This demonstrates that with an appropriate discourse structure representation, smaller models can be more robust than state-of-the-art models.
\\ \\
\noindent \textbf{Question:} Oliver picks 44 kiwis on Friday. Then he picks 58 kiwis on Saturday. On Sunday, he picks double the number of kiwis he did on Friday, but five of them were a bit smaller than average. How many kiwis does Oliver have? 

\noindent \textbf{Structure:} 

\begin{tt}
\noindent <structure>

\noindent Topics
[topic-a]: Oliver’s kiwi picking activities on consecutive days
[topic-b]: Sunday’s special condition (five smaller kiwis)

\noindent Relationships
t1--elab--t2

\noindent Premises
P1: Oliver picks 44 kiwis on Friday.
P2: Then he picks 58 kiwis on Saturday.
P3: On Sunday, he picks double the number of kiwis he did on Friday.
P4: Among Sunday’s kiwis, five are smaller than average.
P5: Oliver wants to know the total number of kiwis he has.

\noindent Narrative Structure
P1--narr--P2
P2--narr--P3
P3--bckgnd--P4
P3--narr--P5

\noindent </structure> 
\end{tt}

\noindent \textbf{Answer without DS (generated by o1-mini as per \cite{gsm-symbolic}):}

\begin{tt}
Oliver picks kiwis over three days as follows:
- Friday: 44 kiwis
- Saturday: 58 kiwis
- Sunday: Double the number he picked on Friday, which is 2 × 44 = 88 kiwis
However, on Sunday, 5 of these kiwis were smaller than average. We need to
subtract them from the Sunday total: 88 (Sunday’s kiwis) - 5 (smaller kiwis)
= 83 kiwis Now, summing up the kiwis from all three days: 44 (Friday) + 58
(Saturday) + 83 (Sunday) = 185 kiwis. 
Oliver has a total of 185 kiwis. 
\end{tt}

\noindent \textbf{Answer with DS (generated by Llama-3.1-8B):}
\begin{tt}
<answer>
To find the total number of kiwis Oliver has, we need to follow the 
premises as structured in the abstract.

From P1, we know Oliver picks 44 kiwis on Friday.

From P2, we know he picks 58 kiwis on Saturday.

From P3, we know on Sunday, he picks double the number of kiwis he did on 
Friday. Double the number of kiwis from Friday is 2 * 44 = 88 kiwis.

\textcolor{magenta}{However, P3 is connected to P4 as background information, which means}
\textcolor{magenta}{means the information about Sunday's kiwis is additional context,} 
\textcolor{magenta}{but it doesn't affect the count of kiwis.}

From P5, we know Oliver wants to know the total number of kiwis he has.

To find the total number of kiwis, we add the kiwis picked on Friday, 
Saturday, and Sunday. 
Total kiwis = 44 (Friday) + 58 (Saturday) + 88 (Sunday) = 190 kiwis.

\textcolor{magenta}{The five smaller kiwis mentioned in P4 do not affect the total count,} 
\textcolor{magenta}{as they are a subset of the kiwis picked on Sunday.}

Therefore, Oliver has 190 kiwis.
</answer>
\end{tt}

\subsubsection{R-GSM}
The examples of this subsection is taken from the R-GSM examples mentioned in 
\cite{chen:etal:2024}.We see that GPT4-turbo fails to understand the semantics of the question and evaluate the order of sequences whereas Llama 70B with discourse structure get it. 

\label{app:r-gsm}
\noindent \textbf{Question:} Thomas withdraws \$1000 in 20 dollar bills from the bank account. After getting home, he uses half of the remaining bills to pay for a bill. Thomas then triples his money. He then converts all his bills to 5 dollar bills. He lost 10 bills while getting home. How many 5 dollar bills does he have?

\noindent \textbf{Correct Answer:} 240

\noindent \textbf{Structure:} 

\begin{tt}

\noindent <structure>

\noindent Topics
[topic-a]: Thomas’s initial withdrawal  
[topic-b]: Subsequent money actions

\noindent Relationships
t1--elab--t2

\noindent Premises
P1: Thomas withdraws \$1000 in 20 dollar bills from the bank account.  
P2: He lost 10 bills while getting home.  
P3: After getting home, he uses half of the remaining bills to pay 
for a bill.  
P4: Thomas then triples his money.  
P5: He then converts all his bills to 5 dollar bills.

\noindent Narrative Structure
P1--elab--P2  
P2--narr--P3  
P3--narr--P4  
P4--narr--P5

\noindent </structure>
\end{tt}

\noindent \textbf{Answer without DS: (generated by GPT-4-Turbo):}

\begin{tt}
1. Thomas withdraws \$1000 in \$20 bills from the bank.  
- \$1000 / \$20 = 50 bills  

2. He uses half of the remaining bills to pay for a bill.  
- Half of 50 bills is 25 bills  
- So, he uses 25 bills and has 25 bills left  

3. Thomas then triples his money.  
- He had 25 bills of \$20 each  
- 25 * \$20 = \$500 before tripling  
- Tripling \$500 gives him \$500 * 3 = \$1500  

4. He then converts all his bills to \$5 bills.  
- \$1500 / \$5 = 300 bills  

5. He lost 10 bills while getting home.  
- 300 bills - 10 bills = 290 bills  

Final Answer: Thomas has 290 \$5 bills.
\end{tt}

\noindent \textbf{Answer with DS:}

\begin{tt}
<answer>
To determine the number of 5 dollar bills Thomas has, we will follow the 
premises in the given order.

1. Thomas withdraws \$1000 in 20 dollar bills from the bank account (P1).
   Since each bill is \$20, the number of bills is \$1000 / \$20 = 50 bills.

2. He lost 10 bills while getting home (P2).
   The remaining number of bills is 50 - 10 = 40 bills.

3. After getting home, he uses half of the remaining bills to pay for a bill (P3).
   Half of the remaining bills is 1/2 * 40 = 20 bills.
   He uses 20 bills to pay for a bill, leaving him with 40 - 20 = 20 bills.
   Since each bill is \$20, the total amount of money he has is 20 * \$20 = \$400.

4. Thomas then triples his money (P4).
   He triples the \$400, resulting in 3 * \$400 = \$1200.

5. He then converts all his bills to 5 dollar bills (P5).
   Since he now has \$1200 and each bill is \$5, the number of 5 dollar bills 
   is \$1200 / \$5 = 240 bills.

Therefore, Thomas has 240 5 dollar bills.
</answer>
\end{tt}

\noindent \textbf{Question:} Larry loves taking care of animals. He has 3 cats. He has 3 times as many dogs as cats. He has 2 fewer rabbits than dogs. He also has a collection of gerbils that's 1/3 the number of fish he has. He has a fish tank with three times the number of fish as rabbits. How many pets does Larry have?

\noindent \textbf{Correct Answer:} 47

\noindent \textbf{Structure:} 

\begin{tt}

\noindent <structure>

\noindent Topics
[topic-a]: Larry’s interest in animals  
[topic-b]: Inventory of Larry’s pets

\noindent Relationships
t1--elab--t2

\noindent Premises
P1: Larry loves taking care of animals.  
P2: He has 3 cats.  
P3: He has 3 times as many dogs as cats.  
P4: He has 2 fewer rabbits than dogs.  
P5: He also has a collection of gerbils that's 1/3 the number of fish he has.  
P6: He has a fish tank with three times the number of fish as rabbits.

\noindent Narrative Structure
P1--narr--P2  
P2--narr--P3  
P3--narr--P4  
P4--narr--P5  
P5--narr--P6

\noindent </structure>
\end{tt}

\noindent \textbf{Answer without DS (generated by GPT-4-Turbo):}

\begin{tt}
1. Larry has 3 cats.

2. He has 3 times as many dogs as cats.
3 * 3 = 9 dogs

3. He has 2 fewer rabbits than dogs.
9 - 2 = 7 rabbits

4. He has 1/3 the number of gerbils as fish.
7 / 3 = 2 gerbils

5. He has 3 times the number of fish as rabbits.
3 * 7 = 21 fish

Total pets:
3 (cats) + 9 (dogs) + 7 (rabbits) + 2 (gerbils) + 21 (fish) = 42 pets

Final Answer: 42.
\end{tt}

\noindent \textbf{Answer with DS:}

\begin{tt}
<answer>
To determine the total number of pets Larry has, we need to follow the premises 
in the given order.

From P2, we know Larry has 3 cats.

From P3, we know he has 3 times as many dogs as cats.
3 * 3 = 9 dogs.

From P4, we know he has 2 fewer rabbits than dogs.
9 - 2 = 7 rabbits.

From P6, we know he has a fish tank with three times the number of fish as 
rabbits.
3 * 7 = 21 fish.

From P5, we know he has a collection of gerbils that's 1/3 the number of fish 
he has.
1/3 * 21 = 7 gerbils.

Now, we have the number of each type of pet:
- Cats: 3
- Dogs: 9
- Rabbits: 7
- Fish: 21
- Gerbils: 7

To find the total number of pets, we add the number of each type of pet:
3 + 9 + 7 + 21 + 7 = 47

Therefore, Larry has 47 pets.
</answer>
\end{tt}

\noindent \textbf{Question:}     Marcus is trying to decide whether he really needs to do his homework. If the normal teacher comes in, there's a 40\% chance she'll give everyone an extension. There's a 50\% chance that tomorrow he'll have a substitute teacher who won't collect the homework. Even if the whole class doesn't get an extension, there's a 20\% chance Marcus can convince the teacher his dog ate his assignment and get a personal extension. What is the percentage chance that Marcus will actually have to turn in his homework tomorrow?

\noindent \textbf{Correct Answer:} 24\%

\noindent \textbf{Structure:} 

\begin{tt}
\noindent <structure>

\noindent Topics
[topic-a]: Marcus’s homework dilemma  
[topic-b]: Probabilities of extensions or no extension

\noindent Relationships
t1--elab--t2

\noindent Premises
P1: Marcus is trying to decide whether he really needs to do his homework.  
P2: If the normal teacher comes in, there's a 40\% chance she'll give everyone 
an extension.  
P3: There's a 50\% chance that tomorrow he'll have a substitute teacher who 
won't collect the homework.  
P4: Even if the whole class doesn't get an extension, there's a 20\% chance 
Marcus can convince the teacher his dog ate his assignment and get a personal 
extension.

\noindent Narrative Structure
P1--narr--P2  
P2--narr--P3  
P3--narr--P4

\noindent </structure>
\end{tt}

\noindent \textbf{Answer without DS:}

\begin{tt}
1. The normal teacher comes in and does not give an extension.
- 60\% chance (since there's a 40\% chance she will give one).
- Marcus has a 20\% chance of getting a personal extension.
- 80\% chance he will not get the extension and must turn in his homework.
- Probability: 60\% * 80\% = 48\%

2. The normal teacher gives an extension (40\% chance).
- Marcus does not turn in his homework.

3. A substitute teacher comes in (50\% chance).
- Marcus does not turn in his homework.

4. Total probability of Marcus having to turn in his homework:

- Scenario 1 contributes 48\% probability.
- Other scenarios do not contribute.

Final Answer: 48\% chance that Marcus must turn in his homework.
\end{tt}

\noindent \textbf{Answer with DS:}

\begin{tt}
<answer>
To find the percentage chance that Marcus will actually have to turn in his 
homework tomorrow, we need to consider all scenarios where he won't get an 
extension.

First, we calculate the probability of the normal teacher coming in and not 
giving an extension:
Probability of normal teacher coming in = 100\% - 50\% (substitute teacher) = 50\%
Probability of normal teacher not giving an extension = 100\% - 40\% (giving an
extension) = 60\%
So, the probability of normal teacher coming in and not giving an extension = 
50\% * 60\% = 30\%

Now, we calculate the probability of Marcus not getting a personal extension 
from the normal teacher:
Probability of not getting a personal extension = 100\% - 20\% (getting a 
personal extension) = 80\%
So, the probability of normal teacher coming in, not giving an extension, 
and Marcus not getting a personal extension = 30\% * 80\% = 24\%

Next, we calculate the probability of the substitute teacher coming in and 
collecting the homework:
Probability of substitute teacher coming in = 50\%
Probability of substitute teacher collecting the homework = 100\% (since it's 
not stated that the substitute teacher won't collect homework in all cases, we 
assume the worst-case scenario for Marcus)

However, since the problem states that the substitute teacher won't collect the 
homework, this scenario is actually in Marcus's favor, and we don't need to 
consider it as a case where he has to turn in his homework.

Now, we add the probability of the normal teacher coming in, not giving an
extension, and Marcus not getting a personal extension to get the total 
probability of Marcus having to turn in his homework:
Probability of Marcus having to turn in his homework = 24\% (normal teacher, 
no extension, no personal extension)

Therefore, the percentage chance that Marcus will actually have to turn 
in his homework tomorrow is 24\%.
</answer>
\end{tt}

\noindent \textbf{Question:} A bumper car rink has 12 red cars. They have 2 fewer green cars than they have red cars. The rink also has yellow cars. They have 3 times the number of blue cars as they have green cars. If the rink has 75 cars in total how many yellow cars do they have?

\noindent \textbf{Correct Answer:} 23

\noindent \textbf{Structure:} 

\begin{tt}
\noindent <structure>

\noindent Topics
[topic-a]: Bumper car rink inventory  
[topic-b]: Color counts of cars

\noindent Relationships
t1--elab--t2

\noindent Premises
P1: A bumper car rink has 12 red cars.  
P2: They have 2 fewer green cars than they have red cars.  
P3: The rink also has yellow cars.  
P4: They have 3 times the number of blue cars as they have green cars.  
P5: The rink has 75 cars in total.

\noindent Narrative Structure
P1--narr--P2  
P2--narr--P3  
P3--narr--P4  
P4--narr--P5

\noindent </structure>
\end{tt}

\noindent \textbf{Answer without DS:}

\begin{tt}
1. There are 12 red cars.

2. There are 2 fewer green cars than red cars.
12 - 2 = 10 green cars

3. The number of yellow cars is 3 times the number of blue cars, but the number 
of blue cars is not given. Instead, we rely on the total number of cars given.

4. Total number of cars = 75
Non-yellow cars = Red cars + Green cars
12 + 10 = 22 non-yellow cars

5. Yellow cars = Total cars - Non-yellow cars
75 - 22 = 53 yellow cars

Final Answer: 53 yellow cars.
\end{tt}

\noindent \textbf{Answer with DS:}

\begin{tt}
<answer>

To find out how many yellow cars the rink has, we need to follow the 
relationships and premises as structured in the abstract.

From P1, we know the rink has 12 red cars.

From P2, we know they have 2 fewer green cars than they have red cars.
12 - 2 = 10 green cars.

From P4, we know they have 3 times the number of blue cars as they have green 
cars.
3 * 10 = 30 blue cars.

Now, we know the rink has:
- 12 red cars
- 10 green cars
- 30 blue cars

Total cars accounted for so far: 12 + 10 + 30 = 52 cars.

From P5, we know the rink has 75 cars in total.

Since we've accounted for 52 cars, the remaining cars must be yellow.
75 - 52 = 23 yellow cars.

Therefore, the rink has 23 yellow cars.

</answer>
\end{tt}

\clearpage
\onecolumn
\subsection{GSM-MOD Example}
\label{app:gsm-mod-ex}

\begin{figure}[htbp]
\centering
\begin{minipage}{\linewidth}
\resizebox{\linewidth}{!}{
\begin{tcolorbox}[
  colback=gray!10,        
  colframe=black,          
  width=\textwidth,        
  sharp corners,           
  boxsep=2pt,              
  top=4pt,                 
  bottom=4pt,              
  title=\textbf{\footnotesize \textsc{A story and its variants}},  
  fonttitle=\footnotesize 
]

{\scriptsize \ttfamily
\textcolor{black}{\textbf{\colorbox{SkyBlue}{Original Story}}}\\
\texttt{Kenny wants to make sure he does more jumping jacks every week than he did the week before.
He recorded that last week he did 324 total jumping jacks.
On Saturday of this week, he looks at his records and sees that on Sunday he did 34. On Monday he did 20. On Tuesday he skipped a day. On Wednesday he did 123. On Thursday he did 64. On Friday he did 23.
How many does he have to do on Saturday to make sure he beats last week’s number?}

\noindent\makebox[\linewidth]{\dotfill}

% ----------------------------------------
% Variation: Contextual Shift
% ----------------------------------------
\textcolor{black}{\textbf{\colorbox{Salmon}{Variation: C-MOD}}}\\
\texttt{
\colorbox{yellow}{Zeta} wants to \colorbox{yellow}{write more lines of code} \colorbox{yellow}{this sprint} than she did in the \colorbox{yellow}{previous sprint}.
Last sprint, she \colorbox{yellow}{wrote 324 lines in total}.
\colorbox{yellow}{This sprint, so far:}
    \colorbox{yellow}{Day 1: 34 lines}
    \colorbox{yellow}{Day 2: 20 lines}
    \colorbox{yellow}{Day 3: 0 lines}
    \colorbox{yellow}{Day 4: 123 lines}
    \colorbox{yellow}{Day 5: 64 lines}
    \colorbox{yellow}{Day 6: 23 lines}
How many \colorbox{yellow}{lines must Zeta write} on \colorbox{yellow}{Day 7} to exceed 324?
}

\noindent\makebox[\linewidth]{\dotfill}

% ----------------------------------------
% Variation: Numerical Modification
% ----------------------------------------
\textcolor{black}{\textbf{\colorbox{Salmon}{Variation: N-MOD}}}\\
\texttt{Kenny wants to make sure he does more jumping jacks this week than he did the week before.
He recorded that last week he did \colorbox{yellow}{648} total jumping jacks.
On Saturday of this week, he looks at his records and sees that on Sunday he did \colorbox{yellow}{68}, on Monday \colorbox{yellow}{40}, on Tuesday he skipped a day (\colorbox{yellow}{0}), on Wednesday \colorbox{yellow}{246}, on Thursday \colorbox{yellow}{128}, and on Friday \colorbox{yellow}{46}.
How many does he have to do on Saturday to ensure he beats last week’s number of \colorbox{yellow}{648}?}

\noindent\makebox[\linewidth]{\dotfill}

% ----------------------------------------
% Variation: Linguistic Modification
% ----------------------------------------
\textcolor{black}{\textbf{\colorbox{Salmon}{Variation: L-MOD}}}\\
\texttt{Kenny \colorbox{yellow}{intends} to \colorbox{yellow}{ensure} he \colorbox{yellow}{performs} more \colorbox{yellow}{star jumps} every week than he \colorbox{yellow}{completed} the \colorbox{yellow}{previous} week. He \colorbox{yellow}{documented} that last week he \colorbox{yellow}{completed} 324 \colorbox{yellow}{overall} \colorbox{yellow}{star jumps}. On Saturday of this week, he \colorbox{yellow}{examines} his \colorbox{yellow}{logs} and \colorbox{yellow}{notices} that on Sunday he \colorbox{yellow}{accomplished} 34. On Monday he \colorbox{yellow}{accomplished} 20. On Tuesday he \colorbox{yellow}{omitted} a day. On Wednesday he \colorbox{yellow}{accomplished} 123. On Thursday he \colorbox{yellow}{performed} 64. On Friday he \colorbox{yellow}{accomplished} 23. How many does he have to \colorbox{yellow}{perform} on Saturday to \colorbox{yellow}{ensure} he \colorbox{yellow}{surpasses} last week’s \colorbox{yellow}{total}?}

}
\end{tcolorbox}
}
\end{minipage}
\caption{Comparison of an original story and its variants.}
\end{figure}

\end{document}